\documentclass[10pt,twocolumn,letterpaper]{article}

\usepackage{cvpr}              
\usepackage{colortbl}  
\usepackage{bbding} 
\usepackage{multirow} 
\usepackage{bbm} 
\definecolor{cvprblue}{rgb}{0.21,0.49,0.74}
\usepackage[pagebackref,breaklinks,colorlinks,allcolors=cvprblue]{hyperref}

\def\paperID{*****} 
\def\confName{CVPR}
\def\confYear{2026}

\title{AdaTok:Adaptive Token Compression with Object-Aware Representations for Efficient Multimodal LLMs}

\author{Xinliang Zhang$^{a,b,c}$, Lei Zhu$^{a,b,c}$, Hangzhou He$^{a,b,c}$, Shuang Zeng$^{a,b,c}$, Ourui Fu$^{a,b,c}$, \\ Jiakui Hu$^{a,b,c}$, Zhengjian Yao$^{a,b,c}$, Yanye Lu$^{a,b,c,*}$ \and $^a$Institute of Medical Technology, Peking University Health Science Center \\
$^b$Department of Biomedical Engineering, Peking University \\
$^c$National Biomedical Imaging Center, Peking University \\
Beijing, China\\
{\tt\small {zhangxinliang}@stu.pku.edu.cn, yanye.lu@pku.edu.cn}
}

\begin{document}
\maketitle
\begin{abstract}
Multimodal Large Language Models (MLLMs) have demonstrated substantial value in unified text-image understanding and reasoning, primarily by converting images into sequences of patch-level tokens that align with their architectural paradigm. However, patch-level tokenization leads to a quadratic growth in image tokens, burdening MLLMs’ understanding and reasoning with enormous computation and memory. Additionally, the traditional patch-wise scanning tokenization workflow misaligns with the human vision cognition system, further leading to hallucination and computational redundancy. To address this issue, we propose an object-level token merging strategy for \textbf{Ada}ptive \textbf{Tok}en compression, revealing the consistency with human vision system. The experiments are conducted on multiple comprehensive benchmarks, which show that our approach averagely, utilizes only 10\% tokens while achieving almost 96\% of the vanilla model’s performance. More extensive experimental results in comparison with relevant works demonstrate the superiority of our method in balancing compression ratio and performance. Our code is provided in the supplementary material.
\end{abstract}    
\section{Introduction}
\label{sec:intro}

\begin{figure}[ht]
    \centering
    \includegraphics[width=0.9\linewidth]{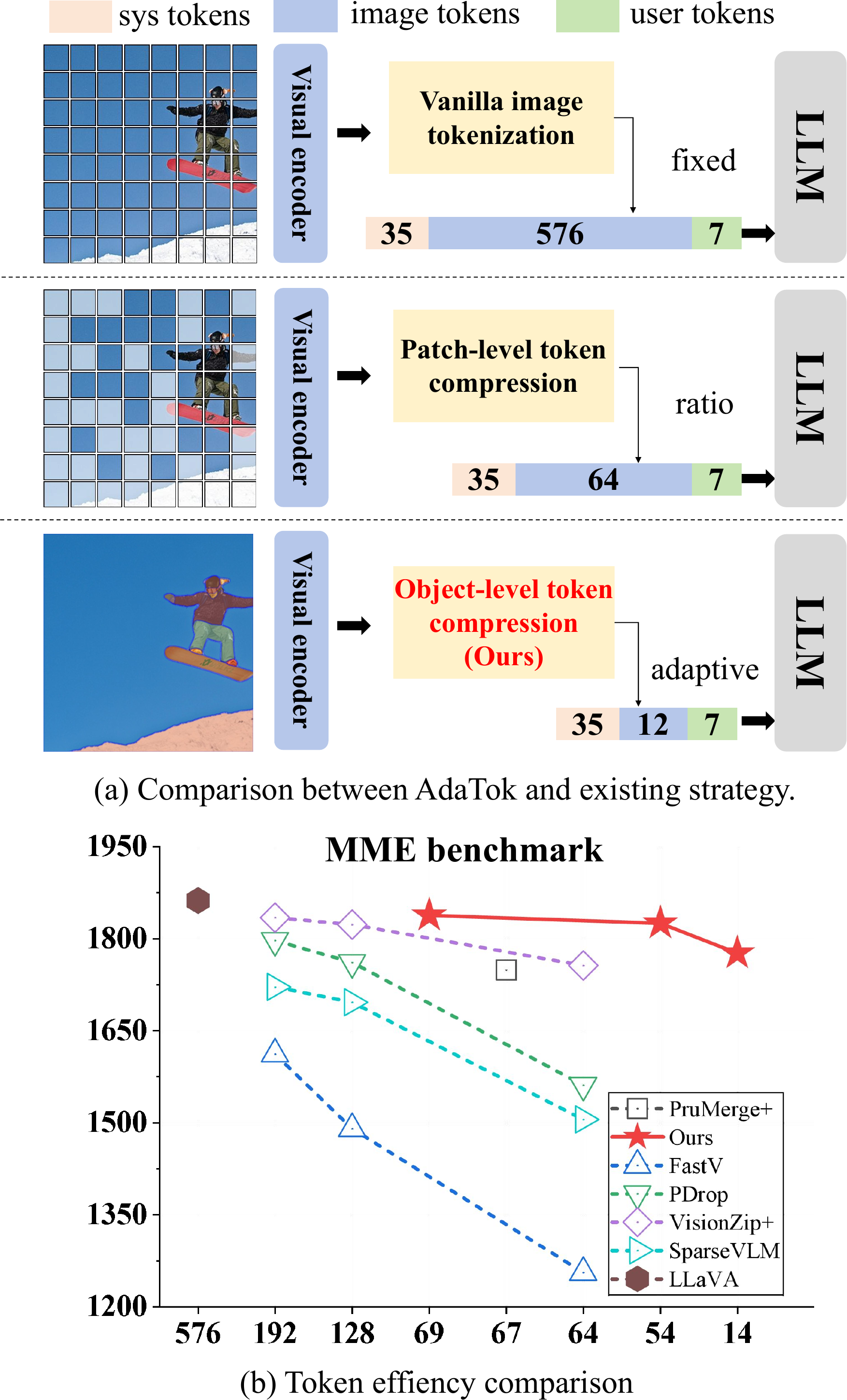} 
    \caption{(a) The main idea of AdaTok. Our object-level image token compression approach inherently outperforms patch-level compression-based methods in terms of the balance on compression ratio and accuracy. (b) The x-axis is token numbers, y-axis is the total score on MME benchmark. }
    \label{fig:mainidea}
\end{figure}

Benefiting from the rapid advances of large language models (LLMs)~\cite{openai2023gpt,bai2023qwen,touvron2023llama,liu2024deepseek}, multimodal large language models (MLLMs)~\cite{lin2023video,li2023llava,liu2024llavanext,dai2023instructblip,li2023blip,chen2023shikra} have become a major research focus in the computer vision community. They have demonstrated remarkable progress across a wide range of vision-language understanding and reasoning tasks~\cite{liu2023llava,liu2024world,liu2023improvedllava,zhu2024beyond,lin2023video,li2023llava}, pushing general artificial intelligence closer to real-world applications.
Mainstream MLLMs typically adopt sequential visual representations extracted by a text-aligned visual encoder (e.g., CLIP-ViT~\cite{radford2021learning}), which are then projected into the text feature space through a lightweight multilayer perceptron. The resulting projected embeddings are referred to as image tokens. However, this patch-by-patch tokenization scheme is inherently inefficient, as it treats all regions equally regardless of semantic importance. Consequently, large models waste substantial computation on redundant tokens during decoding. To this end, visual token compression has become essential for mitigating computational redundancy in MLLMs~\cite{chen2024image,lin2025boosting,xing2024pyramiddrop,yang2025visionzip,shang2024llava,huang2024dynamic}.

To address this issue, several works rank image tokens by their attention scores and discard those with low attention weights according to a predefined compression ratio~\cite{chen2024image,lin2025boosting,xing2024pyramiddrop}, as presented in Figure~\ref{fig:mainidea}(a). However, since the number of objects varies across images, fixed-ratio compression inevitably removes tokens corresponding to valid objects, leading to information loss. This not only limits the model’s ability to fully exploit visual cues but also increases the risk of hallucination during inference. Moreover, such fixed-ratio strategies exhibit poor adaptivity: some objects receive redundant tokens, while others are underrepresented. Consequently, they fail to allocate tokens dynamically according to object distribution and remain constrained by rigid compression ratios.
In addition, several works introduce token fusion strategies~\cite{shang2024llava,yang2025visionzip} to alleviate information loss. Nevertheless, these methods still perform clustering at the patch level without explicitly incorporating object-level semantics. Furthermore, their cluster center computation depends on the similarity between the [CLS] token and patch features, where the overly global nature of the [CLS] representation tends to introduce noisy and object-irrelevant tokens, ultimately degrading model performance~\cite{darcet2023vision}.
Motivated by these limitations, we investigate a key question:
\textit{\textbf{Can image tokens be adaptively compressed based on object-level semantics while preserving the complete visual content as faithfully as possible?}}

To this end, we abandon traditional patch-level token compression and shift the compression granularity to the object level. As illustrated in Figure~\ref{fig:mainidea}(b), we compress the image tokens by merging the visual representations that belong to the same object. This design brings three key advantages. (1) \textbf{Efficiency}: Representing an image with a compact set of object-level tokens is more efficient and aligns better with human visual perception, while patch-wise scanning introduces substantial redundancy. (2) \textbf{Information completeness}: Object-level representations preserve semantic integrity without discarding informative regions, effectively reducing hallucination risks during inference compared with token-dropping methods. (3) \textbf{Adaptability}: As the number of objects varies across images, object-level compression naturally adapts to image content without depending on a fixed compression ratio~\cite{chen2024image,xing2024pyramiddrop,yang2025visionzip}, achieving a better trade-off between compression efficiency and task performance. Benefiting from the above advantages, AdaTok demonstrates significant superiority in token utilization efficiency as presented in Figure~\ref{fig:mainidea}(b).
The main contributions of this paper can be concluded as follows:
\begin{itemize}
    \item An object-aware strategy named AdaTok is proposed for adaptive token compression, which balances the global semantic information and adaptability, where most existing works mainly focus on patch-level representation using much more tokens with non-adaptability.
    \item Our proposed object-aware strategy can be easily integrated into current MLLMs, enhancing the image tokens utilization and inference efficiency, without any internal modifications to the model architecture. 
    \item Extensive experiments on multiple benchmarks are conducted, which demonstrate the efficiency and generalization of our method.

\end{itemize}


\section{Related work}
\label{sec:related_work}

\begin{figure*}[ht]
    \includegraphics[width=\linewidth]{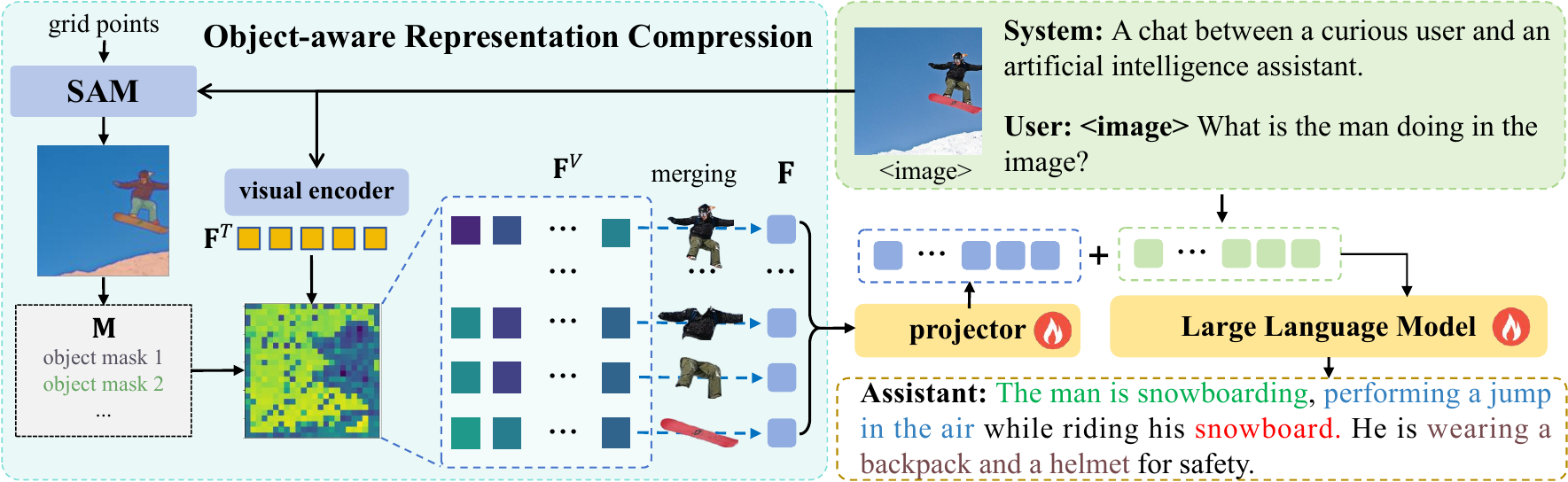}
    \caption{The overview of our object-aware representation merging strategy. We implement our strategy based on LLaVA architecture.}
    \label{fig:overview}
\end{figure*}
\subsection{Efficient foundation models}
Large language models (LLMs) have demonstrated strong capabilities in text tasks such as reading comprehension and question answering~\cite{bai2023qwen,openai2023gpt,touvron2023llama,liu2024deepseek}. Multimodal large language models (MLLMs) further extended these capabilities to more complex image-text cross-modal understanding and reasoning tasks~\cite{li2023llava,liu2023visual,li2023blip,dai2023instructblip,zhang2024llava,wu2025f,zhang2024omg}. In the early exploration stage of multimodal large models, some methods have focused on reducing the number of tokens in the visual backbone network~\cite{bolya2022tome,2023PuMer,kong2112spvit,vasu2025fastvlm}, while others have attempted to design dedicated projectors to enable efficient injection of visual information~\cite{li2023blip,li2025tokenpacker}. Among the open-sourced alternatives, LLaVA~\cite{liu2023improvedllava} offers a concise yet effective pipeline that utilizes only a ViT backbone network and an MLP to facilitate the injection of visual information. In this study, we adopt LLaVA-v1.5 as our research subject.

\subsection{Dropping based token compression}
 To address the issue of image token redundancy, token dropping based methods involve designing token-disposal strategies based on attention scores~\cite{chen2024image,lin2025boosting,xing2024pyramiddrop,zhang2024sparsevlm}. By operating on a specific transformer block inside the LLM backbone, FastV~\cite{chen2023shikra} ranks the tokens according to the attention score and drops the tokens that are below a particular threshold. VTW~\cite{lin2025boosting} proposed to withdraw all the image tokens at a certain layer of the LLM backbone, where the layer selection will directly affect the performance of the model. PyDrop~\cite{xing2024pyramiddrop} operates with different dropping ratios according to the layer index of the LLM backbone. SparseVLM~\cite{zhang2024sparsevlm} devised a token recycling mechanism to reuse tokens from the deleted pools, thereby offsetting the model degradation caused by token reduction. However, dropping tokens will inevitably lose the informative tokens and hinder further trade-offs between compression ratio and accuracy.
 \subsection{Merging based token compression}
 Merging based methods~\cite{shang2024llava,yang2025visionzip} focus on developing token clustering and fusion strategies centered around the ViT's [CLS] token pretrained by CLIP~\cite{radford2021learning}. Prumerge~\cite{shang2024llava} adopted the [CLS] token as the global information reference to cluster the image tokens via K-means~\cite{likas2003global} and implement the weighted average pooling on each cluster for token merging. VisionZip~\cite{yang2025visionzip} preserved the image tokens that have a high attention score to the [CLS] token as the dominant tokens, while merging other inferior image tokens via their similarity to each other. Specifically, DynamicLLaVA~\cite{huang2024dynamic} additionally trained a transformer block with an MLP to select the image tokens in a certain layer of the LLM backbone. However, the [CLS] token of CLIP-ViT may include artifacts due to the absence of registers~\cite{darcet2023vision}.

\section{Method}
\label{sec:method}

In this chapter, we first review the workflow of MLLMs and the ambition of visual token compression. Subsequently, we elaborate on the technical details of the proposed object-aware representation merging strategy for token compression. Finally, we will demonstrate the advantages of our strategy over patch-level compression.

\subsection{Problem Definition}
MLLMs follow an autoregressive paradigm, where pre-filled tokens and newly generated tokens are combined to serve as input for the next decoding step. In the pre-filling phase, the image is tokenized to a sequential embedding $\mathbf{X}^I$ through a visual encoder and a visual projector, while the text is mapped to $\mathbf{X}^T$ via a text tokenizer. Then, $\mathbf{X}^T$ and $\mathbf{X}^I$ are concatenated together as the initial input $\mathbf{X}$, and then fed into the LLM backbone for autoregressive inference. Usually, the length of $\mathbf{X}^I$ is typically much larger than the length of $\mathbf{X}^T$, accounting for the majority of the computational burden. With an LLM backbone stacked by $L$ transformer layers, denoting the transformer decoder as $\mathcal{D}$, the computational cost $\mathcal{T}$ is positively correlated with the square of the length of $\mathbf{X}$, which can be expressed using the following formula:
\begin{align}
        & \mathcal{T} ((\mathcal{D}_{1:k-1}(\mathbf{X}_1))\circ(\mathcal{D}_{k:L}(\mathbf{X}_k))) \\ \notag
        & \propto (k-1) |\mathbf{X}_1|^2 + (L-k)|\mathbf{X}_k|^2.
\end{align}
where $\mathbf{X}_k$ represents the input tokens of $k$-th transformer decoder layer. Assuming we have a compression function that can compress the length of $\mathbf{X}_1$ with a ratio $r\in[0,1]$, the benefit can be expressed as:
\begin{equation}
     \nabla\mathcal{T} \propto (L-k)(1-r^2).
     \label{eq:benefit}
\end{equation}
The detailed derivation is provided in the appendix. From Eq.~\ref{eq:benefit}, the benefits brought by compressing tokens depend on two variables: the position $k$ where the compression function takes effect and the compression ratio $r$. The earlier the compression function takes effect (i.e., the smaller $k$ is), the greater the compression degree and the smaller $r$ becomes.

Given the $k$-1$^{th}$ layer's tokens as the input, the goal of the token compression field is to design a compression function $\mathcal{C}$, which can be described as:
\begin{equation}
    \mathbf{X}_k = \mathcal{C}(\mathbf{X}_{k-1};r,\delta),
    \label{eq:compression-function}
\end{equation}
where $\delta$ is a prior corresponding to the specific strategy.

For token dropping based methods~\cite{chen2024image,lin2025boosting,xing2024pyramiddrop,zhang2024sparsevlm}, $\mathcal{C}$ is devised as a ranking and dropping strategy with $k>1$, and $\delta$ is acquired from the patch-level attention score, which contains little cues about the objects. For this strategy, due to the lack of object-related priors, the selection of $k$ and $r$ requires careful consideration, and it is prone to information loss, which in turn leads to the occurrence of hallucinations. For token merging based methods~\cite{shang2024llava,yang2025visionzip}, $\mathcal{C}$ is devised as a clustering strategy with $k=1$, adopting the [CLS] token of CLIP-pretrained ViT as the prior $\delta$ to acquire the cluster center. Though the [CLS] token aggregates the global information of the whole image, it also involves high noise tokens, resulting in misalignment with the correct objects.

In comparison, our strategy takes effect in the input stage with $k$= 1, and $\mathcal{C}$ is an object-aware representation compression function, $\delta$ is the object masks, whose number further decides the compression ratio $r$. It is worth noting that, as the number of objects in each image is different, $r$ adaptively varies according to the image.
\subsection{Object-aware representation compression}
An overview of our method is presented in Figure~\ref{fig:overview}. We extract the object masks of the input image in the object-level merging module, and the object masks will be applied to the visual features from the CLIP-ViT to guide the representation merging, acquiring the compressed representation of the input image. Then the compressed representation will be sent to the visual projector and generate the compressed image tokens, which will be concatenated with system tokens and user prompt tokens together and fed into the large language model for auto-regressive generation.

We utilize the SAM~\cite{kirillov2023segment} to extract all the object masks. In detail, we generate a grid point set with $p$ points per side as the positive prompt input. The SAM will output a series of masks for the object at each point and assign a corresponding confidence score. We filter out the masks whose confidence score is below $\sigma$, which is set to 0.8 by default. Denoting the image representation before the projection as $\mathbf{F}^T$, we reshape these serialized image embeddings back into square format and upsample them to the original image resolution, getting the spatially aligned visual embeddings $\mathbf{F}^{V}\in\{f ^V_i \in \mathbb{R}^{1\times d}|i=1,..., H\times W\}$, where $H$ and $W$ represents the height and width of the image. Denoting $\mathbf{M}=\{m_i \in \mathbbm{1}^{H\times W} |i=1,...,k\}$ as the masks for all objects in the entire image, for each object mask, we implement the average merging on $\mathbf{F}^{V}$:
\begin{equation}
    f_i =\frac{1}{\|m_i\|_1} \sum m_i \cdot \mathbf{F}^V ,
\end{equation}
producing the compressed image embeddings $\mathbf{F}=\{ f_i\in \mathbb{R}^{1\times d}|i=1,...,k\}$, where the length of $\mathbf{F}$ equals to the number of the objects. The compressed features will be sent to the projector, which is a two-layer MLP, to generate the compressed image tokens. Then the compressed image tokens will be concatenated with system tokens and user tokens, and fed to LLM for autoregressive generation.

\subsection{Model training}
As each compressed token represents a complete object, there is a gap between the information of the compressed tokens and the LLM, which further degrades the model. After all, the original LLM is trained with patch-level tokens, which only represent a part of the object. To bridge this gap, we adopt the instruction tuning data of LLaVA-1.5~\cite{liu2023improvedllava} to finetune the projector and LLM backbone with LoRA~\cite{hu2022lora} for 1 epoch. The batchsize per device is set to 14 due to the GPU memory limitation, other settings for tuning are kept the same with LLaVA-1.5 stage 2. We adopt the SAM~\cite{kirillov2023segment} with ViT-H backbone to generate the object-level masks, with default points per side ``p'' set to 32.


\subsection{Advantages over patch-level compression}
The lack of object-related cues is an inherent flaw of patch-level methods. This flaw forces these approaches to rely on additional priors and precise designs to compress visual tokens, which prevents them from achieving a small compression ratio $r$. Furthermore, our method does not involve any modifications to the model structure or calls to internal model parameters, which helps reduce the pipeline costs of product development. Besides, the object-level merging phase operates entirely in the prefilling phase, which means the whole process of object-level merging can be deployed to the terminal device, thereby reducing the computational pressure on the server side. This is highly useful in scenarios with limited communication bandwidth, such as maritime communication, satellite Internet, and communication in remote areas. We will calculate the bandwidth required for communication in Section~\ref{sec:bandwidth}.
\begin{figure*}[ht]
\centering
\includegraphics[width=0.95\linewidth]{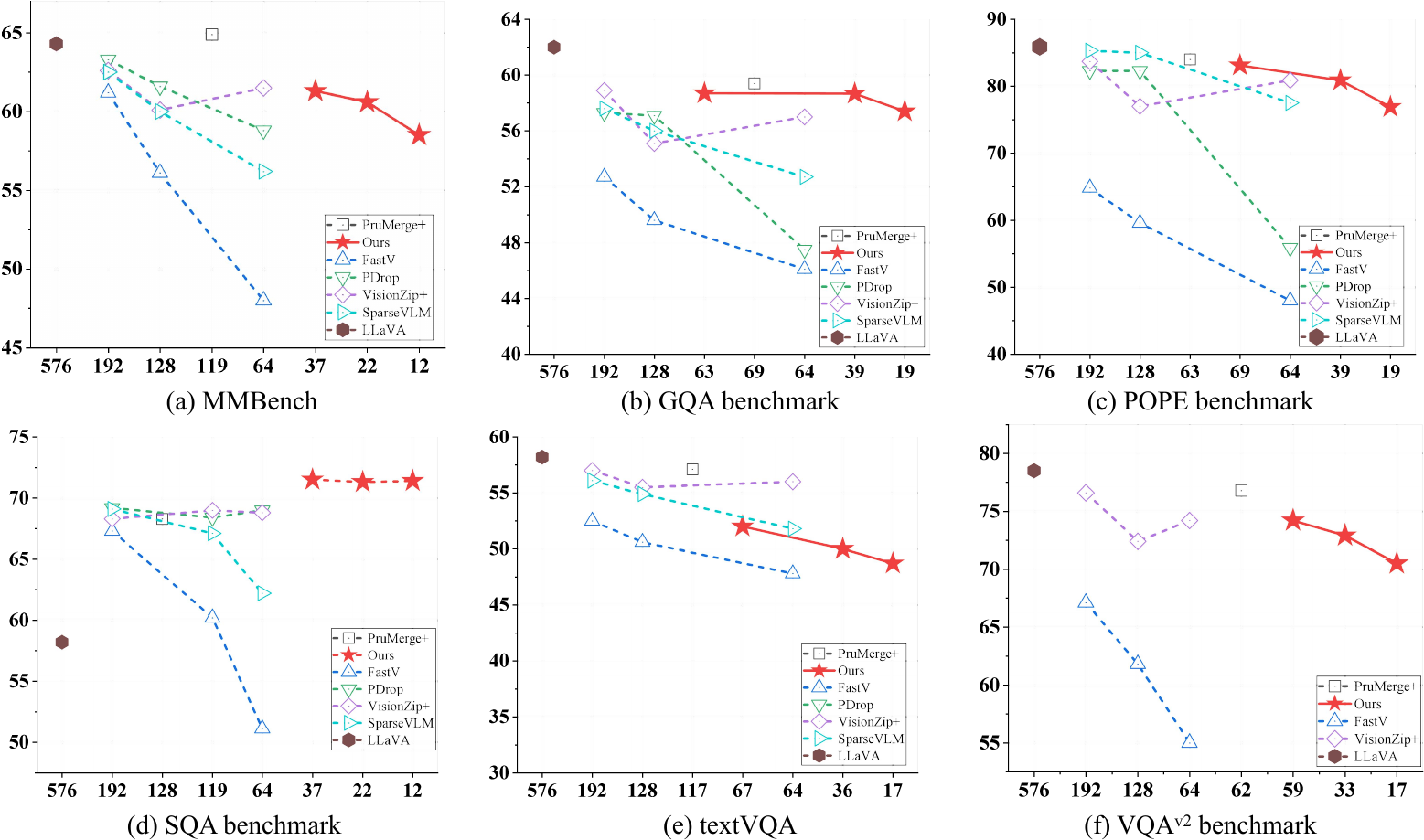}
\caption{Token efficiency comparison on different benchmarks. Our proposed object-aware token merging strategy requires fewer tokens to maintain comparable performance, showcasing its inherent advantage over patch-level based compression strategies.}
\label{fig:TokensScores}
\end{figure*}
\section{Experiments}\label{sec:experiments}
\begin{table*}[ht]
\small
\tabcolsep=0.15cm
\centering
\begin{tabular}{lccccccccccc}
\toprule
Method            & Publish & Tokens$\downarrow$ & LLM       & Adaptive & VQAv²$\uparrow$ & SQA$\uparrow$  & textVQA$\uparrow$ & POPE$\uparrow$ & MME$\uparrow$  & MMB$\uparrow$  & GQA$\uparrow$  \\
\midrule
LLaVA-1.5         & 23'NIPS & 576    & Vicuna-7B & \XSolidBrush        & 78.5  & 66.8 & 58.2 & 85.9 & 1862 & 64.3 & 62   \\
\hline
\multicolumn{12}{c}{Patch-level token dropping based methods}                                                 \\
\hline
\rowcolor[HTML]{EFEFEF} 
PyDrop\cite{xing2024pyramiddrop}             & 25'CVRP & 64     & Vicuna-7B & \XSolidBrush        & -     & 69.0 & 50.6 & 55.9 & 1561 & 58.8 & 47.5 \\
PyDrop\cite{xing2024pyramiddrop}             & 25'CVRP & 128    & Vicuna-7B & \XSolidBrush        & -     & 68.4 & 56.6 & 82.3 & 1761 & 61.6 & 57.1 \\
PyDrop\cite{xing2024pyramiddrop}             & 25'CVRP & 192    & Vicuna-7B & \XSolidBrush        & -     & 69.2 & 56.5 & 82.3 & 1797 & 63.3 & 57.3 \\
PyDrop+\cite{xing2024pyramiddrop}            & 25'CVRP & 270    & Vicuna-7B & \XSolidBrush        & -     & 71.0 & 58.5 & 86.0 & -    & 66.1 & 61.9 \\
\rowcolor[HTML]{EFEFEF} 
SparseVLM\cite{zhang2024sparsevlm}         & 25'ICML & 64     & Vicuna-7B & \XSolidBrush        & -     & 62.2 & 51.8 & 77.5 & 1505 & 56.2 & 52.7 \\
SparseVLM\cite{zhang2024sparsevlm}         & 25'ICML & 128    & Vicuna-7B & \XSolidBrush        & -     & 67.1 & 54.9 & 85.0 & 1696 & 60.0 & 56.0 \\
SparseVLM\cite{zhang2024sparsevlm}         & 25'ICML & 192    & Vicuna-7B & \XSolidBrush        & -     & 69.1 & 56.1 & 85.3 & 1721 & 62.5 & 57.6 \\
\rowcolor[HTML]{EFEFEF} 
FastV\cite{chen2024image}             & 24'ECCV & 64     & Vicuna-7B & \XSolidBrush        & 55.0  & 51.1 & 47.8 & 48.0 & 1256 & 48.0 & 46.1 \\
FastV\cite{chen2024image}             & 24'ECCV & 128    & Vicuna-7B & \XSolidBrush        & 61.8  & 60.2 & 50.6 & 59.6 & 1490 & 56.1 & 49.6 \\
FastV\cite{chen2024image}             & 24'ECCV & 192    & Vicuna-7B & \XSolidBrush        & 67.1  & 67.3 & 52.5 & 64.8 & 1612 & 61.2 & 52.7 \\
\hline
\multicolumn{12}{c}{Patch-level token merging based methods}                                                  \\
\hline
\rowcolor[HTML]{EFEFEF} 
VisionZip+\cite{yang2025visionzip}        & 25'CVPR & 64     & Vicuna-7B & \XSolidBrush        & 74.2  & 68.8 & \textbf{56.0} & 80.9 & 1756 & 61.5 & 57.0 \\
VisionZip+\cite{yang2025visionzip}        & 25'CVPR & 128    & Vicuna-7B & \XSolidBrush        & 72.4  & 69.0 & 55.5 & 77.0 & 1823 & 60.1 & 55.1 \\
VisionZip+\cite{yang2025visionzip}        & 25'CVPR & 192    & Vicuna-7B & \XSolidBrush        & 76.6  & 68.3 & 57.0 & 83.7 & 1834 & 62.6 & 58.9 \\
\rowcolor[HTML]{EFEFEF} 
Prumerge\cite{shang2024llava}          & 25'ICCV & 54     & Vicuna-7B & \Checkmark        & 72.0  & 68.5 & 56.0 & 76.3 &      & 60.9 & 55.9 \\
Prumerge+\cite{shang2024llava}         & 25'ICCV & 89     & Vicuna-7B & \Checkmark        & 76.8  & 68.3 & 57.1 & 84.0 & 1749 & 64.9 & -    \\
DynLLaVA-I\cite{huang2024dynamic}    & 25'ICLR & 115    & Vicuna-7B & \Checkmark        & 78.0  & 69.1 & 57.0 & 85.0 & -    & 65.4 & 61.4 \\
DynLLaVAI-I/T\cite{huang2024dynamic} & 25'ICLR & 115    & Vicuna-7B & \Checkmark        & 77.9  & 68.6 & 56.5 & 85.9 & -    & 64.1 & 61.3 \\
\hline
\multicolumn{12}{c}{Object-level token merging based methods}                                                 \\
\hline
Ours(p=8)         & -       & 15     & Vicuna-7B & \Checkmark        & 70.5  & 71.4 & 48.7 & 76.9 & 1776 & 58.5 & 57.4 \\
Ours(p=16)        & -       & 30     & Vicuna-7B & \Checkmark        & 72.9  & 71.3 & 50.0 & 80.9 & 1783 & 60.6 & 58.7 \\
\rowcolor[HTML]{EFEFEF} 
Ours(p=32)        & -       & 53     & Vicuna-7B & \Checkmark        & \textbf{74.2}  & \textbf{71.5} & 52.2 & \textbf{83.3} & \textbf{1819} & \textbf{61.3} & \textbf{58.6} \\

\end{tabular}
\caption{Performance of our method on LLaVA-1.5 with Vicuna-7B as the LLM backbone, $\uparrow$ the higher the better, $\downarrow$ the lower the better. The rows corresponding to the method that requires the minimum number of tokens are highlighted, with the best result \textbf{bolded}.}
\label{tab:main_results}
\end{table*}

\begin{table*}
\centering
\small
\begin{tabular}{lccccccccc} 
\toprule
Method         & Publish  & Token$\downarrow$ & GQA$\uparrow$   & MMB$\uparrow$   & ~ MME$\uparrow$ & POPE$\uparrow$  & SQA$\uparrow$   & VQAv2$\uparrow$ & VQAText$\uparrow$  \\ 
\hline
LLaVA-Next-7B  & arXiv    & 2880   & 64.2~ & 67.4~ & 1851~ & 86.5~ & 70.1~ & 81.8~ & 64.9~    \\ 
\hline
\multicolumn{10}{c}{Patch-level dropping based methods}                                       \\ 
\hline
\rowcolor[HTML]{EFEFEF} 
FastV~\cite{chen2024image}          & 24'ECCV  & 320    & 55.9~ & 61.6~ & 1661~ & 71.7~ & 62.8~ & 71.9~ & 55.7~    \\
\rowcolor[HTML]{EFEFEF} 
PyDrop~\cite{xing2024pyramiddrop}          & 25'CVPR  & 320    & 56.4~ & 63.4~ & 1663~ & 77.6~ & 67.5~ & 73.5~ & 54.4~    \\
\rowcolor[HTML]{EFEFEF} 
FasterVLM~\cite{vasu2025fastvlm}      & 25'ICCV  & 320    & 56.9~ & 61.6~ & 1701~ & 83.6~ & 66.5~ & 74.0~ & 56.5~    \\
\rowcolor[HTML]{EFEFEF} 
HiRED~\cite{arif2025hired}          & 25'AAAI  & 320    & 59.3~ & 64.2~ & 1690~ & 83.3~ & 66.7~ & 75.7~ & 58.8~    \\
\rowcolor[HTML]{EFEFEF} 
DART~\cite{wen2025stop}           & 25'EMNLP & 320    & 61.7~ & \textbf{65.3}~ & 1710~ & \textbf{84.1}~ & 68.4~ & 79.1~ & 58.7~    \\
\rowcolor[HTML]{EFEFEF} 
HoloV~\cite{zou2025don}          & 25'NIPS  & 320    & \textbf{61.7}~ & 65.3~ & \textbf{1738}~ & 83.9~ & 68.9~ & \textbf{79.5}~ & \textbf{58.7}~    \\
SparseVLM~\cite{zhang2024sparsevlm}      & 25'ICML  & 640    & 60.3  & 65.7  & 1772  & -     & 67.7  & 77.1  & 57.8     \\
SparseVLM~\cite{zhang2024sparsevlm}      & 25'ICML  & 320    & 57.7  & 64.3  & 1694  & -     & 67.3  & 73.4  & 55.9     \\
\rowcolor[HTML]{EFEFEF} 
SparseVLM~\cite{zhang2024sparsevlm}      & 25'ICML  & 160    & 51.2  & 63.1  & 1542  & -     & 67.5  & 66.3  & 46.4     \\ 
\hline
\multicolumn{10}{c}{Patch-level merging based methods}                                        \\ 
\hline
\rowcolor[HTML]{EFEFEF} 
Prumerge+\cite{shang2024llava} & 25'ICCV  & 320~ & 53.6~ & 61.3~ & 1534~ & 60.8~ & 66.4~ & 69.7~ & 50.6~    \\
VisionZip+\cite{yang2025visionzip}     & 25'CVPR  & 640    & 62.4  & 65.9  & 1778  & -     & 67.9  & 79.9  & 60.8     \\
VisionZip+\cite{yang2025visionzip}     & 25'CVPR  & 320    & 61.0    & 64.4  & 1770  & -     & 67.5  & 78.4  & 59.3     \\
\rowcolor[HTML]{EFEFEF} 
VisionZip+\cite{yang2025visionzip}     & 25'CVPR  & 160    & 58.2  & 63.9  & 1699  & -     & 67.5  & 75.6  & 57.3     \\ 
\hline
\multicolumn{10}{c}{Object-level token merging based methods}                                 \\ 
\hline
Ours(p=8)  & - & 43~ & 54.5~ & 50.0~ & 1449~ & 76.3~ & 68.7~ & 69.0 & 45.9~  \\
Ours(p=16) & - & 83  & 55.8~ & 54.6~ & 1484~ & 81.1~ & \textbf{68.9}~ & 71.6  & 49.2~  \\
\rowcolor[HTML]{EFEFEF} 
Ours(p=32) & - & 140 & 56.1~ & 54.8~ & 1495~ & 83.6~ & 68.7~ & 72.9     & 52.4~  \\
\bottomrule
\end{tabular}
\caption{Performance of our method on LLaVA-Next with Vicuna-7B as the LLM backbone, $\uparrow$ the higher the better, $\downarrow$ the lower the better.$\uparrow$ the higher the better. The rows corresponding to the method that requires the minimum number of tokens are highlighted, with the best result \textbf{bolded}. }
\label{tab:llava-next}

\end{table*}
\subsection{Benchmarks}

We conducted experiments on multiple benchmark datasets, including: MME~\cite{zhang2024mme} is a systematical benchmark that evaluates the comprehensive performance of models in terms of perception and cognition; POPE~\cite{Li-hallucination-2023} is designed to assess the hallucinations of MLLMs; TextVQA~\cite{singh2019towards} evaluates the ability to recognize text through OCR of MLLMs; SQA~\cite{lu2022learn} is short for scientific question and answering, whic assesses the scientific reasoning and knowledge alignment of MLLMs; GQA~\cite{hudson2019gqa} measures the scene understanding ability of MLLMs; MMBench~\cite{MMBench} access the visual-related reasoning and perception ability of the MLLMs; and the traditional vision question answering dataset VQAv2~\cite{balanced_vqa_v2}.

\subsection{Main results}

We present the results of our AdaTok in Table~\ref{tab:main_results}, and include the recent image token compression methods for comparison. Patch-level token dropping based methods include PyDrop~\cite{xing2024pyramiddrop}, SparseVLM~\cite{zhang2024sparsevlm}, VisionZip$+$~\cite{yang2025visionzip}, and FastV~\cite{chen2024image}. It has to be noted that these methods need a fixed ratio for image token compression, which cannot adaptively adjust the tokens according to the number of objects in the image. Patch-level token merging based methods include Prumerge~\cite{shang2024llava} and DynLLaVA~\cite{huang2024dynamic}. Prumerge$+$ enhanced the performance of Prumerge by involving more tokens.

In Table~\ref{tab:main_results}, we highlight the minimum number of tokens required by other methods as well as their results across various benchmarks. Since the number of image tokens in our method corresponds one-to-one with the number of objects in an image, for fair comparison, we increased the parameter p (points per side) to raise the number of sampling points, thereby obtaining more objects, ensuring our token count is as aligned as possible with that of these methods. The conclusion can be drawn from the table that our method achieves better performance on VQAv², SQA, POPE, MME, MMB, and GQA with fewer tokens. Notably, on the POPE benchmark, our method retains 96.9\% of the performance of LLaVA-1.5, while patch-level methods like PyDrop, SparseVLM, and FastV exhibit a significant performance drop. This reflects the inherent flaw of patch-level token dropping based methods, that they are more prone to hallucinations. To achieve a similar performance, VisionZip$+$ needs 192 tokens, while ours just needs 53 tokens. In more extreme cases, we set the points per side to 8, where only 15 tokens are required (\textbf{only 2.6\% tokens}). On the MME benchmark, our method even outperforms PyDrop (128 tokens), SparseVLM (192 tokens), VisionZip+ (64 tokens), FastV (192 tokens), and Prumerge+ (89 tokens), while \textbf{preserving 95.4\% performance} of the vanilla LLaVA-1.5. With respect to the textVQA benchmark, our method faces some challenges due to the inherent limitation of the object-level merging strategy, which will be discussed in Section~\ref{sec:limitation}.
\subsection{The token efficiency comparison.}
\begin{table}[ht]
\small
\tabcolsep=0.1cm
\centering
\begin{tabular}{lccccccc} 
\toprule
Method  & VQAv² & SQA   & VQAT  & POPE   & MME     & MMB   & GQA    \\ 
\hline
\multicolumn{8}{c}{vision tower resolution 336}                     \\ 
\hline
Vanilla & 78.5  & 66.8  & 58.2  & 85.9   & 1862    & 64.3  & 62.0     \\
frozen  & 61.0~ & 67.8~ & 46.6~ & 70.5~  & 1259~   & 42.9~ & 50.9~  \\
S1      & 0.1~  & 37.4~ & 0.3~  & 0.7~   & 0~      & 42.9~ & 0.1~   \\ 
S1+S2   & 73.2~ & 69.7~ & 49.8~ & 82.3~  & 1623~   & 58.8~ & 58.0~  \\
S2      & \textbf{74.2}~ & \textbf{71.5}~ & \textbf{52.2}~ & \textbf{83.3}~  & \textbf{1819}~   & \textbf{61.2}~ & \textbf{58.6}~  \\
\hline
\multicolumn{8}{c}{vision tower resolution 224}                     \\ 
\hline
Vanilla & 76.5 & 70.3~ & 54.2~ & 84.4~  & 1794~   & 62.6~ & 60.5~  \\
frozen  & 63.5~ & 68.4~ & 46.6~ & ~75.1 & 1427~ & 46.7~ & 52.4~  \\
S1      & 0.6~  & 42.2~ & 2.5~  & 0.6~   & 9.0~    & 13.2~ & 0.5~   \\
S1+S2   & 72.7~ & 70.3~ & 49.6~ & 82.2~  & \textbf{1741}~ & 58.6~ & 57.9~  \\
S2      & \textbf{72.9}~ & \textbf{71.3}~ & \textbf{49.8}~ & \textbf{81.6}~  & 1661~ & \textbf{59.5}~ & \textbf{57.9}~  \\
\bottomrule
\end{tabular}
\caption{Ablation study on fine-tuning each component with Vicuna-7B as the language backbone. ``S1'' and ``S2'' denote different visual-instruction tuning stage of LLaVA-v1.5-7B.}
\label{tab:ablation}
\end{table}
It is a consensus in the current field that the higher the compression ratio of image tokens, the more information about the image will be lost. Therefore, we tested the token utilization efficiency of our method and its robustness to the compression ratio of tokens across multiple benchmarks. In Figure~\ref{fig:TokensScores}, we present the relationship between the number of tokens in our method and the metrics across different datasets, with ``p'' set to [8, 16, 32]. Compared with existing patch-level methods, our approach requires fewer tokens to maintain comparable performance, which reflects the inherent advantage of the object-aware strategy. Specifically, in the SQA benchmark, which is a pure scientific reasoning benchmark, our method only requires 6 tokens to achieve satisfactory results. However, our method exhibits suboptimal performance on the textVQA benchmark. This inherent limitation will be discussed in Section~\ref{sec:limitation}.

\section{Discussion}
\subsection{Ablation study}
The main component of our method is the same as the vanilla LLaVA-1.5-7B~\cite{liu2023improvedllava}, including a CLIP-ViT vision tower, a two-layer MLP projector, and a Vicuna-7B language backbone. The original instruction tuning setup of LLaVA-1.5 consists of two phases. In the first stage, the LLM is frozen and only the visual projector is trained. In the second stage, both the LLM and the visual projector are trained simultaneously. Following this setup, we validated our method on different benchmarks with different resolution pretrained visual encoders, and the experimental results are presented in Table~\ref{tab:ablation}. It can be drawn from Table~\ref{tab:ablation} that our object-level merging based method does not require the first step to realign the visual projector. Instead, it only needs to perform the second step tuning on the pre-trained visual projector. Notably, this aspect differs from the patch-level merging method~\cite{shang2024llava,yang2025visionzip}.


\subsection{Generalization and applicability.} 

\begin{table}[ht]
\tabcolsep=0.06cm
\centering
\resizebox{\linewidth}{!}{
\begin{tabular}{lccccccccc} 
\toprule
Model     & Param & VQAv² & SQA   & VQAT  & POPE  & MME   & MMB   & GQA    \\ 
\hline
SAM-H     & 632M  & 74.2~ & 71.5~ & 52.2~ & 83.3~ & 1819~ & 61.2~ & 58.6~  \\
SAM-L     & 308M  & 74.1~ & 71.6~ & 51.9~ & 82.5~ & 1826~ & 60.4~ & 58.9~  \\
SAM-B     & 91M   & 71.5~ & 71.5~ & 50.3~ & 79.5~ & 1746~ & 59.6~ & 57.5~  \\
MobileSAM & 10M   & 72.6~ & 71.5~ & 50.4~ & 80.8~ & 1811~ & 60.8~ & 58.2~  \\
\bottomrule
\end{tabular}}
\caption{Mask quality influence by different visual foundation models. Our method is robust to mask quality and is highly compatible with MobileSAM. }
\label{tab:sam_vit}
\end{table}
We also implement our strategy on the LLaVA-Next architecture to demonstrate its generalization. The results are presented in Table~\ref{tab:llava-next}. In general, our AdaTok inherently utilizes fewer image tokens while maintaining competitive performance. Especially on POPE benchmark, AdaTok outperforms FastV (320 tokens), PyDrop (320 tokens), Prumerge+ (320 tokens), HiRED (320 tokens), reflecting its robustness against hallucinations.

We also experiment with SAM models of other scales to compare the impact of different mask qualities on model performance, including ViT-H, ViT-L, ViT-B, and MobileSAM~\cite{tiny_vit}, as presented in Tables~\ref{tab:sam_vit}. This indicates that our strategy benefits from high-quality object-level masks: the higher the mask quality, the better the model performance. Though the overall impact is not significant, this also reflects the robustness of our strategy. Among them, MobileSAM~\cite{tiny_vit} is sufficiently lightweight and can be deployed on edge devices, further demonstrating the flexibility and applicability of our method.
\begin{figure*}[ht]
    \centering
    \includegraphics[width=\linewidth]{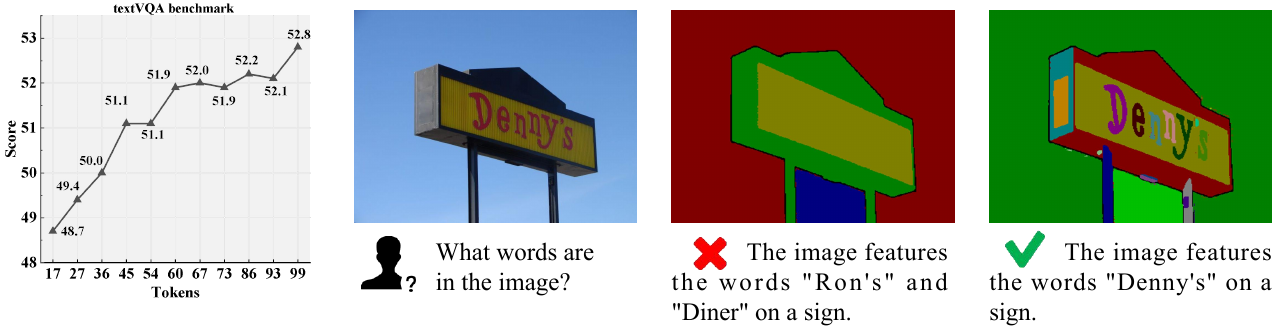}
    \caption{Limitation of AdaTok on OCR task. It is challenging to extract the region of individual letter in the image. To address this issue, it is necessary to increase the number of sampling points to acquire more object masks.}
    \label{fig:OCR_limitation}
\end{figure*}
\begin{figure*}[ht]
\centering
\includegraphics[width=0.95\linewidth]{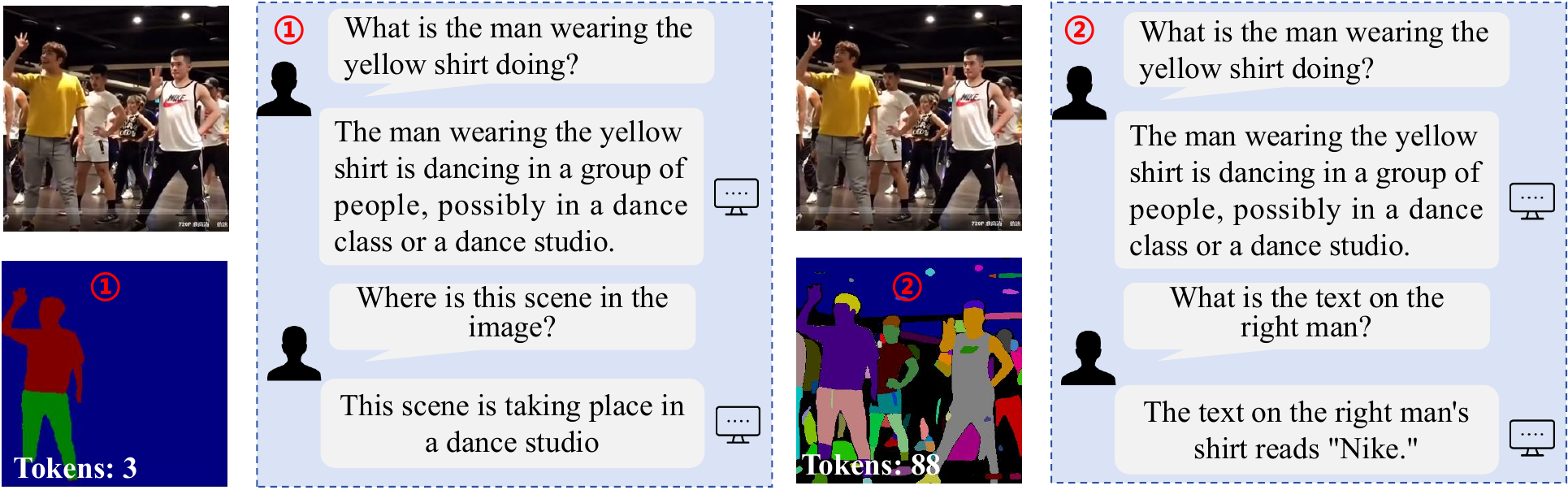}
\caption{The visual question answering results under different tokens. For simple visual question answering tasks such as scene recognition and image captioning, AdaTok only requires the number of relevant key objects to complete the task. For fine-grained recognition tasks such as text recognition, it can also be accomplished by increasing the number of tokens in the image. We can adjust the number of tokens input to AdaTok to flexibly handle tasks of varying complexity. LLaVA-v1.5-7B is adopted here.
 }
    \label{fig:vqa_demo}
\end{figure*}
\subsection{Compression benefit for communication }\label{sec:bandwidth}

\begin{table}[ht]
\centering
\small
\tabcolsep=0.15cm
\begin{tabular}{ccc|ccc} 
\toprule
\multicolumn{1}{l}{Resolution} & \multicolumn{1}{l}{Bandwidth} & Unit & \multicolumn{1}{l}{Tokens} & \multicolumn{1}{l}{Bandwidth} & Unit  \\ 
\hline
224$^2$                            & 147                           & KB/s & 8                          & \textbf{16}                            & KB/s  \\
336$^2$                            & 330.75                        & KB/s & 12                         & \textbf{24}                            & KB/s  \\
480$^2$                            & 675                           & KB/s & 16                         & \textbf{32}                            & KB/s  \\
512$^2$                            & 768                           & KB/s & 32                         & \textbf{64}                            & KB/s  \\
640$^2$                            & 1.17~                         & MB/s & 64                         & \textbf{128}                           & KB/s  \\
768$^2$                            & 1.69~                         & MB/s & 128                        & \textbf{256}                           & KB/s  \\
1024$^2$                           & 3.00~                         & MB/s & 192                        & \textbf{384}                           & KB/s  \\
\bottomrule
\end{tabular}
\caption{The bandwidth requirement of raw image and tokens.}
\label{tab:bandwidth}
\end{table}
It is worth noting that the object-level merging phase operates entirely in the prefilling phase and is completely decoupled from the LLM backbone, which means the whole process of object-level merging can be deployed to the terminal device, thereby reducing the computational pressure on the server side. This is highly useful in scenarios with limited communication bandwidth, such as maritime communication, satellite internet communication, and communication in remote areas.

Taking VQAv² as an example, which is built upon MSCOCO 2014~\cite{lin2014microsoft}, the average image resolution of this benchmark is $640\times 480$. Theoretically, an image with a resolution of 640×480 contains a total of 307,200 pixels. When transmitted in the uint8 format, it requires 921,600 bytes in total, corresponding to a bandwidth of at least 900 KB/s. In contrast, under our default setting of p=32, an average of only 59 tokens (with dimensions 1×1024) are needed on this dataset. When these tokens are transmitted with float16 precision, each token takes 2 bytes, resulting in a total of 120,832 bytes, which is approximately 120 KB/s. Compared with transmitting the original image, the bandwidth requirement is reduced by about 7.5 times after compression using our method. We provide the bandwidth required for natural images at different common resolutions, as well as the bandwidth required for transmitting image tokens after different levels of compression, as shown in Table~\ref{tab:bandwidth}.

\subsection{Flexibility and adaptability}
In Figure~\ref{fig:vqa_demo}, we present the visual question answering results of AdaTok with different numbers of image tokens as the input. We control the number of masks by adjusting the number of sampling points, resulting in two extreme cases of input tokens. It can be observed that, for coarse-grained questions such as scene recognition in case 1 of Figure~\ref{fig:vqa_demo}, the model can generate correct responses even if all objects in the background are merged into a single token. For more fine-grained questions, as long as the relevant regions are segmented, the model's output will not be affected. This means we can adjust the tokens input to AdaTok according to the specific task, enabling flexible adaptation to various complex scenarios. 

\subsection{Limitation of object-level merging}\label{sec:limitation}
Benefiting from SAM's zero-shot transfer capability, our object-level merging strategy works well in most scenarios, balancing compression ratio and accuracy. However, for text recognition tasks, accurately identifying text regions is a significant challenge for SAM, which limits the model's performance. As demonstrated in Figure~\ref{fig:OCR_limitation}, if SAM fails to accurately segment letters, the model will be unable to generate correct responses. To alleviate this issue, it is necessary to increase the number of sampling points to obtain as many letters as possible. However, increasing the points will also lead to over-segmentation, resulting in token redundancy. In the future, we will attempt to use more accurate text recognition models as alternatives to SAM, thereby optimizing the performance on text recognition tasks.


\section{Conclusion}
Visual tokens in multimodal large language models contain significant redundancy, making visual token compression essential for enhancing computational efficiency. In this work, we propose an object-level image token compression strategy that adaptively compresses image tokens based on the number of objects present in each image. Extensive experiments across multiple benchmarks validate the effectiveness of our approach, demonstrating superior performance in achieving a better trade-off between compression ratio and model accuracy.
{
    \small
    \bibliographystyle{ieeenat_fullname}
    \bibliography{main}
}


\end{document}